\documentclass[sigconf]{acmart}
\usepackage{graphicx}
\usepackage{subcaption}
\usepackage{amsmath}
\usepackage{float}

\AtBeginDocument{%
  }

\setcopyright{acmlicensed}
\copyrightyear{2024}
\acmYear{2024}
\acmDOI{XXXXXXX.XXXXXXX}

\acmConference[Conference acronym 'XX]{Make sure to enter the correct
  conference title from your rights confirmation emai}{June 03--05,
  2018}{Woodstock, NY}
\acmISBN{978-1-4503-XXXX-X/18/06}

\begin{document}

\title{Vehicle routing problem using deep reinforcement learning—A case study about truck planning in the industry}

\author{Siliang Lu}
\email{siliang.lu@cn.bosch.com}
\affiliation{%
  \institution{Bosch Center for Artificial Intelligence}
  \city{Shanghai}
  \country{China}
}

\author{Lili Wu}
\email{liliawu@foxmail.com}
\affiliation{%
  \institution{School of Electronic Information
and Electrical Engineering}
  \city{Shanghai}
  \country{China}}

\author{Dan Hu}
\email{hudan.must@gmail.com}
\affiliation{%
  \institution{School of Computer Science, Macau University of Science and Technology}
  \city{Macau}
  \country{China}
}






\renewcommand{\shortauthors}{Siliang Lu et al.}

\begin{abstract}
  As an important component of the supply chain industry, transportation has experienced rapid development in the past decade with the assistance of digital platforms and intelligent algorithms. Within the field of transportation research, Vehicle Routing Problem (VRP) has remained a persistent and enduring challenge. In the realm of management science, experts, and scholars from both the industrial and academic sectors have continuously explored optimization models and algorithms to effectively address routing problems, from the classical Traveling Salesman Problem to the more general Vehicle Routing Problem. These models and algorithms are applied in real-world industrial scenarios to achieve cost optimization and reduce carbon footprints. However, due to the complexity of real-world problems, numerous specific constraints are often added, and challenges such as information opacity, uncertainty, and irrational human behavior may arise. Therefore, deploying and optimizing mathematical models for VRP in practical scenarios while maintaining optimal results poses numerous challenges. This paper discusses and provides solutions for three different logistic use cases involving external truck network design. Through these industrial case study, the paper introduces how deep reinforcement learning-based vehicle routing optimization has been implemented. As a result, it can be observed that the routes optimized by reinforcement learning agent have over 10\% total cost compared to baseline results. Furthermore, the paper proposes that in future research, DRL algorithms for vehicle routing problems could be generalized into more variations of VRP.
\end{abstract}

\ccsdesc[500]{Do Not Use This Code~Generate the Correct Terms for Your Paper}
\ccsdesc[300]{Do Not Use This Code~Generate the Correct Terms for Your Paper}
\ccsdesc{Do Not Use This Code~Generate the Correct Terms for Your Paper}
\ccsdesc[100]{Do Not Use This Code~Generate the Correct Terms for Your Paper}

\keywords{Vehicle routing problems, Deep reinforcement learning, Transformer network}
\begin{teaserfigure}
\centering
  \includegraphics[width=\textwidth ,height=0.3\textwidth]{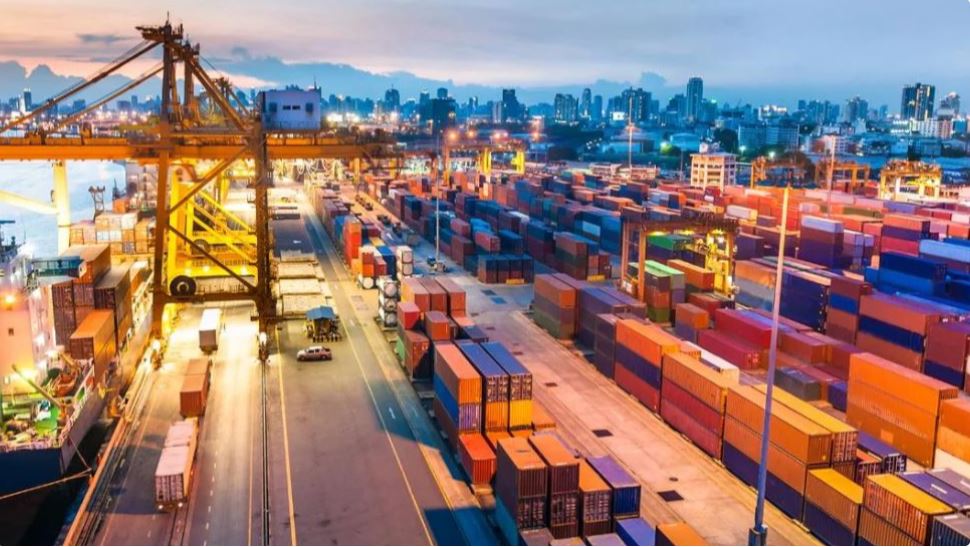}
  \caption{Transportation in the industry}
  \Description{Vehicle Routing Problem (VRP)}
  \label{fig:teaser}
\end{teaserfigure}


\maketitle

\section{Introduction}
Vehicle Routing Problem (VRP) typically refers to organizing and dispatching a fleet of vehicles to service a series of delivery and pickup stations by planning appropriate routes. The goal is for the vehicles to sequentially visit these stations/points to achieve certain objectives (like minimizing the total mileage, reducing overall transportation costs, ensuring vehicles arrive at certain times, or minimizing the number of vehicles used) while satisfying specified constraints (such as delivery and pickup time constraints, vehicle capacity limits, driving distance limits, driving time limits, etc.).

In addition, VRP is also closely linked to corporate logistics models, such as pickup and delivery modes, heterogeneous vehicle types, multi-warehouse mode, and multi-trip mode. To solve VRP, a common approach is to develop an optimization model. However, due to complexity of real-world problems, there may also be issues during real deployment with uncertainties. Therefore, it is challenging to guarantee optimal results for the deployment in real-world scenarios.

In terms of VRP, there are mainly three different solution categories, which are mixed integer programming, heuristic optimization and learning-based optimization. For mixed integer \newline programming (MIP)[\citep{bouzid2017integration}, \citet{singh2021branch},\citet{bach2016branch}], despite such algorithms are able to achieve optimal solutions for small-scale problems within a reasonable cost by establishing strict mathematical models. However, for large-scale problem, as the number of decision variables increases and the constraints of real-world problems become more complex, there is a risk that no feasible solution may be found. Therefore, heuristic optimization methods [\citet{christiaens2020slack},\citet{clarke1964scheduling},\citet{gillett1974heuristic}] such as simulated annealing have emerged to guarantee a near-optimal solution by intuitive or experiential constructs and providing a feasible solution for a combinatorial optimization problem within an acceptable cost. Even if heuristic optimization can provide a feasible solution for a specific instance, it has difficulty in generalizing the solution to other instances. Therefore, with rapid development of learning-based methods for combinatorial optimization, reinforcement learning has illustrated advantages over heuristic optimization regarding generalization and inference efficiency. Hence, this paper mainly aims to discuss about a deployable model-free transformer-based policy network[\citet{li2021deep}] and its application on truck planning in the industry. 

As a well-known problem in management science, this paper examines the heterogeneous capacitated VRP (HCVRP) with real industrial applications using deep reinforcement learning. The paper has three contributions below:
begin{\begin{itemize}
    \item This paper extends the variation of VRP by incorporating a hybrid modes for both less-than-truckload (LTL) transportation and HCVRP for real scenarios, including constraints and objective functions.
    \item A Transformer-based deep reinforcement learning algorithm has achieved the objectives of cost reduction (e.g., over 10\% cost reduction) and improved load factor (e.g., no more than 25\% empty load) compared to baselines.
    \item It provides a deployable solution for combinatorial optimization problem with learning-based algorithms.
\end{itemize}}



\section{Methodology}
\label{sec:others}
\subsection{Modeling}
\label{sec:footnotes}
The set of locations/points is defined as \( X = \{x^i\}_{i=0}^n \). Since the starting point (depot) can be different in this case, each location can serve as a starting point \( x^0 \), which is represented by its absolute coordinates and a demand of 0. The remaining locations describe the coordinates of the starting and ending points for each order. The expression for the set of remaining locations is \( X' = X \setminus \{x^0\} \). Each of these remaining locations is expressed as \( (s^i, d^i) \), where \( s^i \) contains the coordinates of both starting and ending points, and \( d^i \) includes the demand for each order.

Furthermore,  the set of different vehicle types is defined as \( V = \{v^i\}_{i=1}^m \), where \( v^i \) represents the maximum capacity (i.e. weight capacity, volume capacity or quantity capacity) of each vehicle type \( \{Q^i\} \) and m is the number of vehicle types. All the aforementioned variables are non-negative integers. Although in the current real-world scenario, different orders may have pickups at the starting point and deliveries at the ending point, the worst-case scenario is considered here, where all orders need to be unloaded at the last stop of the route. Therefore, the capacity of the vehicle chosen for a route must be no less than the total number of all orders on the route. Besides capacity-related constraints, time-related constraints can added according to specific real-world problems. For instance, each vehicle on a route shall not drive longer than 24h. 
Lastly, the cost function for the VRP in the paper should be:
\begin{equation} \label{cost_function}
\min \sum_{v \in V} \sum_{i \in X} \sum_{j \in X}  \text{fixed cost} + \left(D(x^i, x^j) \times \text{unit cost} \right) \times y_{i,j}
\end{equation}
Here, the pair of fixed cost value and unit cost vary over vehicle type, which are given inputs of the cost function. Meanwhile, the fixed cost and unit cost is calculated with the pricing model provided by domain experts according to vendor data. In addition, because \( x^i \) represents the coordinates of both start and end points for an order, the distance \( D(x^i, x^j) \) represents distance between \( x^i \) and \( x^j \) . Additionally, \( y_{i,j} \) is a binary variable representing whether these two adjacent locations exist in the route.

\subsection{States}
\label{sec:figures}
In MDP problem, each state is defined as \( s_t = (V_t, X_t) \) where \( V_t = \{v_t^1, v_t^2, \ldots, v_t^m\} = \{(o_t^1, T_t^1, G_t^1), (o_t^2, T_t^2, G_t^2), (o_t^3, T_t^3, G_t^3), \ldots\} \), where \( o_t^i \) represents the remaining capacity of the corresponding selected vehicle, \( T_t^i \) represents the accumulated time spent at timestep \( t \), and \( G_t^i = (g_0^i, g_1^i, \ldots) \) represents the set of  coordinates passed through at timestep \( t \), i.e., the information of orders already delivered. To maintain dimension consistency, the number of points/locations included in \( G_t^i \) for each vehicle type \( v_t^i \) at timestep \( t \) is the same. 

Assuming that if only vehicle type \( v_t^1 \) has a new location/point added at timestep \( t \), then the other vehicle types for corresponding routes will repeat adding the last point that has already been included in the corresponding route. Therefore, in the initial state,  $ V_0 = \{(Q^1, 0, \{0\}), (Q^2, 0, \{0\}), \ldots\}$ is the state of vehicles departing from the depot.

At the same time, in each state, the point/location state is \( X_t = \{x_t^1, x_t^2, \ldots, x_t^m\} = \{(s_t^0, d_t^0), (s_t^1, d_t^1), (s_t^2, d_t^2), \ldots, (s_t^m, d_t^m)\} \), where \( s_t^i \) represents the coordinate location and \( d_t^i \) represents the order volume transported at timestep \( t \). 

\subsection{Actions}
\label{sec:tables}
In this paper, the action space is divided into two parts: the first part is to select the optimal vehicle, and the second part is to design the route with the lowest transportation cost. Therefore, the action \( a_t \) at timestep \( t \), denoted as \( a_t \in A \), can be expressed as \( (v_t^i, x_t^i) \), meaning that vehicle \( v_i \) will pass through the starting point or a specific relative coordinate \( x_i \) at timestep \( t \). When sampling the action space, only one action is sampled at each timestep \( t \), including coordinates of a location/point and one vehicle. Furthermore, for actions that do not satisfy the constraints (such as remaining capacity less than 0), real-time updating masks can be established to ensure that the action obtained from the sampling space satisfies the constraints.

\subsection{Reward}
\label{sec:equations}
According to real-world scenario, the objective function of the VRP is to minimize the delivery cost. Therefore, the reward function is designed as the negative of the delivery cost of the truck at the current timestep \( t \), for example, \( R = -\sum_{i=1}^m \sum_{c=1}^C r_t \) where m is the number of vehicles. 

Assuming that at timestep \( t \) and timestep \( t+1 \), \( x^j \) and \( x^k \) are the location/point states at timestep \( t \) and \( t+1 \) respectively, and they are both handled by the same vehicle, then the reward function for time \( t+1 \) can be described as below:

\begin{equation}\label{reward_function}
\begin{aligned}
r_{t+1} = r(s_{t+1}, a_{t+1}) = \{0, \ldots,\text{fixed\_cost} \times \{0, 1\} + D(x^j, x^k) \times \\
\text{unit\_cost}, 0, \ldots\}
\end{aligned}
\end{equation}

For function approximation, since Vehicle Routing Problem (VRP) and its variants are classical graph problems with hard constraints, this neural network primarily adopts Transformer-based architectures to train the policy network.

\section{Case study}
\label{sec:final}
\begin{figure}[ht]
    \begin{center}
    \includegraphics[width=0.4\textwidth]{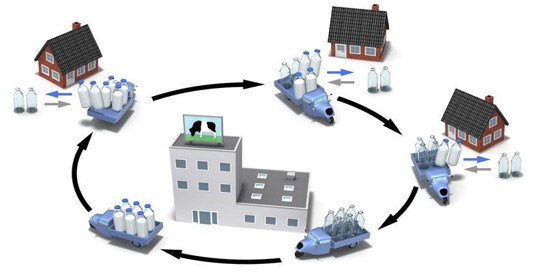}
    \end{center}
    \caption{Illustration about Milk-run}
    \label{fig:example}
\end{figure}

In the context of lean manufacturing in the industry, a supply chain term for the way materials are transported across different factories is called "external milk run". As the name implies, a milk run is a way of delivering milk where many outlets need milk while each needs only a small amount, as shown in Figure \ref{fig:example}. Therefore, using one vehicle for delivery covers various outlets helps cost savings and inventory reduction. Such scenario also describes what external truck routes could look like, especially for manufacturing plants.

In order to apply the proposed learning-based optimization method into real-world, three different external milk-run (EMR) use cases across multiple industrial plants were analyzed as testing datasets. For those three datasets, demands come from multiple manufacturing plants across cities near Shanghai. Moreover, same as training and validation datasets, the vehicle speed is assumed to be 35km/h. Furthermore, with the input features of daily demand data from one point to another point across plants, vehicle type information, geographical locations of the locations, service windows (e.g., from 8 a.m. to 6 p.m.), and service time for loading/unloading service to take (e.g., 0.75 hours), etc., the real use cases can be transformed into mathematical models. 

In terms of constraints for those three use cases, the capacity of each vehicle type in the first use case is represented using the goods quantity, which are 6, 20 , 26, 30, and 44. However, the other two use cases use weight and volume limits.  In detail, the weight capacity of each truck type is 1.9t, 4.75t, 7.6t, 9.5t, 19.0t and the volume capacity of each truck type is 10.368m$^3$, 30.0288m$^3$, 33.5616m$^3$, 40.0896m$^3$, 60.0m$^3$.

Additionally, besides MR, there is another transportation mode called less-than-truck-load transport (LTL) where each supplier individually transports goods from one point to another. Although the transportation cost does not vary over distances traveled and LTL cost is usually higher than milk-run cost, it is still possible LTL is a must option due to customers' considerations or due to lower LTL cost.

Hence, in order to generate different training and validation samples, we strive to closely mimic the three real datasets. Therefore, we use the real value range of each input feature from real use cases to generate more fake training and validation samples. As the demand representation (QTY) in the first use differs from the other two (weight \& volume), Table \ref{tab:dataset_selection1} and Table \ref{tab:dataset_selection2} represents training input features of two different training models, respectively.

\begin{table}[ht]
\centering
\begin{tabular}{|l|l|}
\hline
\textbf{Feature} & \textbf{Range} \\ \hline
Longitude          & 28.000 - 32.000 \\ \hline
Latitude           & 118.000 - 122.000 \\ \hline
{Demand} (QTY)   & 1-44 \\ \hline
Service Time (hour) & 0.75 - 1.5 \\ \hline
\end{tabular}
\caption{Training \& Validation Dataset input feature range for 1st use case}
\label{tab:dataset_selection1}
\end{table}

\begin{table}[ht]
\centering
\begin{tabular}{|l|l|}
\hline
\textbf{Features} & \textbf{Range} \\ \hline
Longitude          & 28.000 - 32.000 \\ \hline
Latitude           & 118.000 - 122.000 \\ \hline
$\text{Demand}_1$ (ton)   & 0.007-17.5 \\ \hline
$\text{Demand}_2$ (m$^3$)   & 0.5-41.4 \\ \hline
Service Time (hour) & 0.75 - 1.5 \\ \hline
\end{tabular}
\caption{Training \& Validation input feature range for 2nd \& 3rd use cases}
\label{tab:dataset_selection2}
\end{table}


Besides decision variables in the classical heterogeneous capacitated VRP model described in section 2, another variable about whether each order is delivered through EMR or through LTL is needed in the case studies. To determine whether the specific order shall be delivered through LTL or through EMR, LTL mode is selected only if the route delivers just a single order and cost of LTL is lower than that of EMR. In addition, in order to generate LTL price for each sample in the training set, we calculate the LTL price based on the real demand's LTL calculation method provided by domain experts. Finally, 10 training samples and 8 validation samples are generated as the training dataset and the validation dataset with each sample containing the demand of 200 orders, respectively.

\begin{figure}[ht]
    \centering
    \begin{subfigure}{0.48\textwidth}
        \centering
        \includegraphics[width=\textwidth]{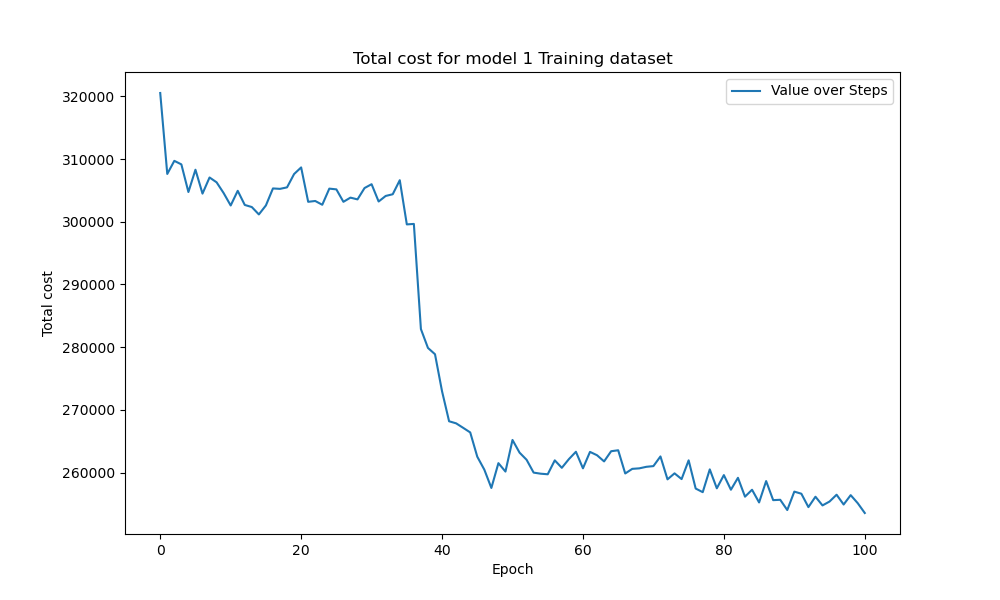}
        \caption{Model1 on training dataset}
        \label{model1_training}
    \end{subfigure}
    \hfill
    \begin{subfigure}{0.48\textwidth}
        \centering
        \includegraphics[width=\textwidth]{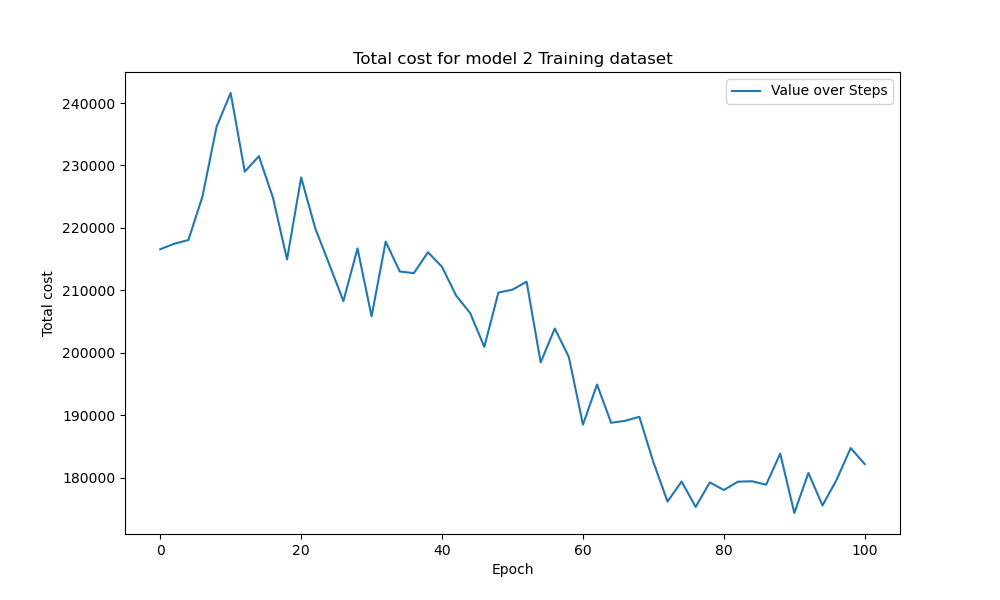}
        \caption{Model2 on training sets}
        \label{model2_training}
    \end{subfigure}
    \caption{Average total costs on the training dataset}
    \label{fig:training}
\end{figure}

\begin{figure}[ht]
    \centering
    \begin{subfigure}{0.48\textwidth}
        \centering
        \includegraphics[width=\textwidth]{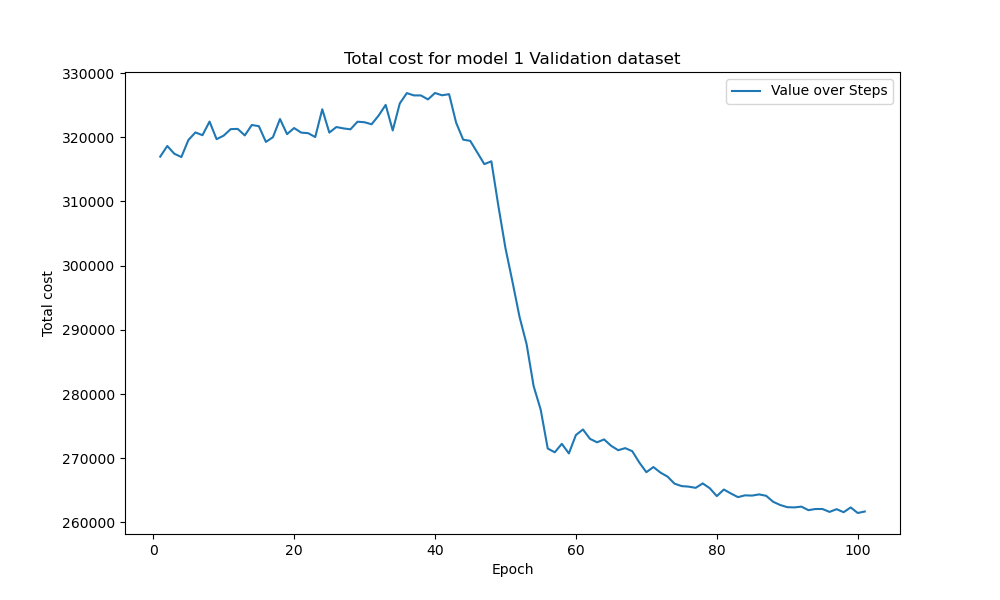}
        \caption{Model1 on validation dataset}
        \label{model1_val_dataset}
    \end{subfigure}
    \hfill
    \begin{subfigure}{0.48\textwidth}
        \centering
        \includegraphics[width=\textwidth]{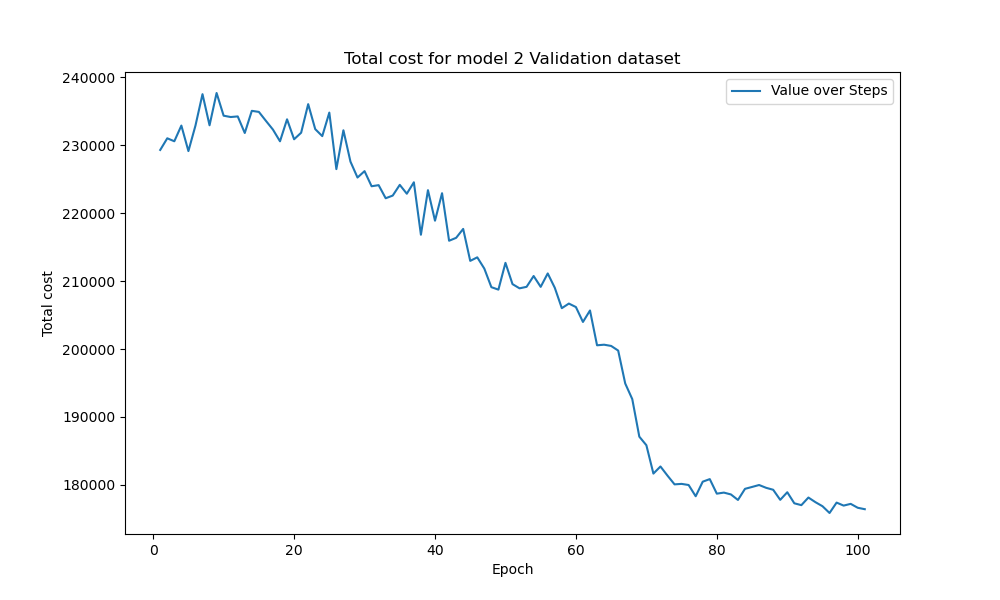}
        \caption{Model2 on validation dataset}
        \label{model2_val_dataset}
    \end{subfigure}
    \caption{Average rewards on the validation dataset}
    \label{fig:validation}
\end{figure}

We used a Quadro RTX 8000 GPU for training two models: the first model was used for the 1st use case, and the second model was used for the 2nd and 3rd use cases. The first model's and the second model's training time were both approximately 5 hours for 100 epochs. Moreover, during the training process, the graph size used by the models was 200, which corresponds to the number of demands. 
During the training process, the performances with the average total transportation costs of model1 and model2 on the training dataset costs are shown in Figure \ref{model1_training} and Figure \ref{model2_training}. As shown in the figures, both cost values has been decreased over the number of epochs. In addition, the performances with the average rewards of model1 and model2 on the validation dataset are shown in Figure \ref{model1_val_dataset} and Figure \ref{model2_val_dataset}. As shown in the figures, as rewards are negative values, both rewards decrease over the number of epochs, which indicates performance improvements during training.

\begin{figure}
    \centering
    \begin{subfigure}{0.4\textwidth}
        \centering
        \includegraphics[width=\textwidth]{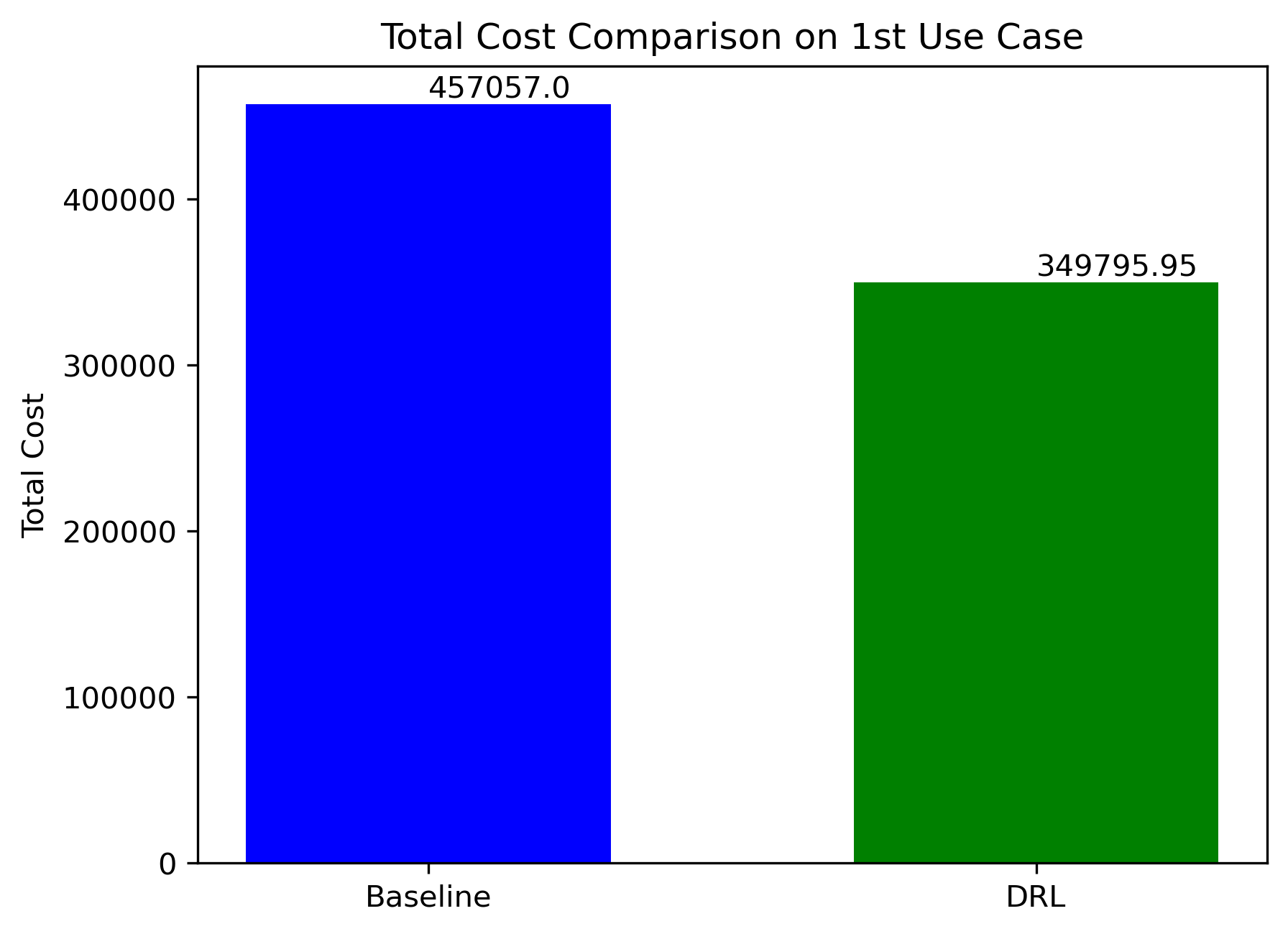}
        \caption{First use case}
        \label{rbcd_res}
    \end{subfigure}
    \hfill
    \begin{subfigure}{0.4\textwidth}
        \centering
        \includegraphics[width=\textwidth]{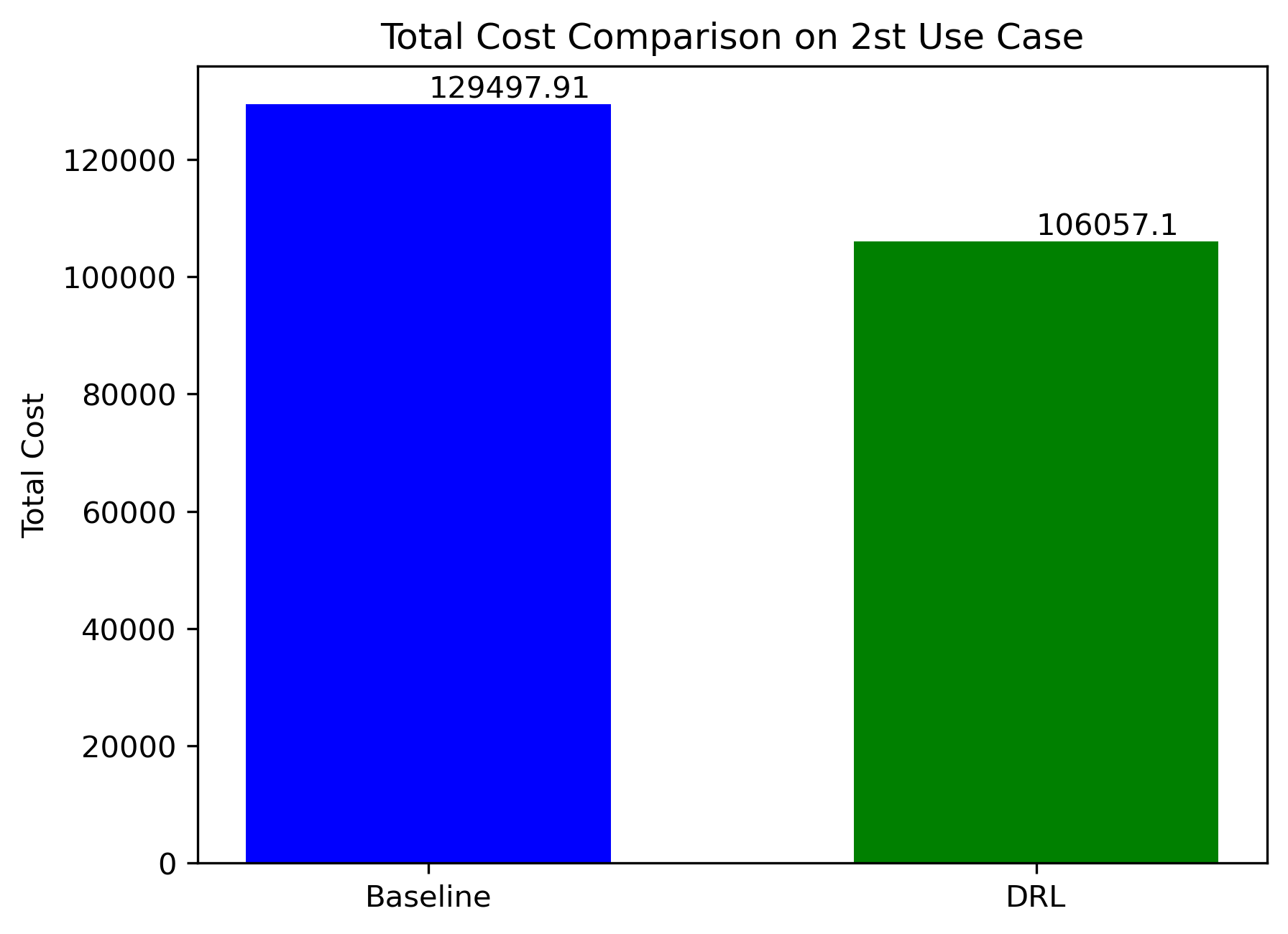}
        \caption{Second use case}
        \label{e5_res}
    \end{subfigure}
    
    \vspace{0.4cm}
    
    \begin{subfigure}{0.4\textwidth}
        \centering
        \includegraphics[width=\textwidth]{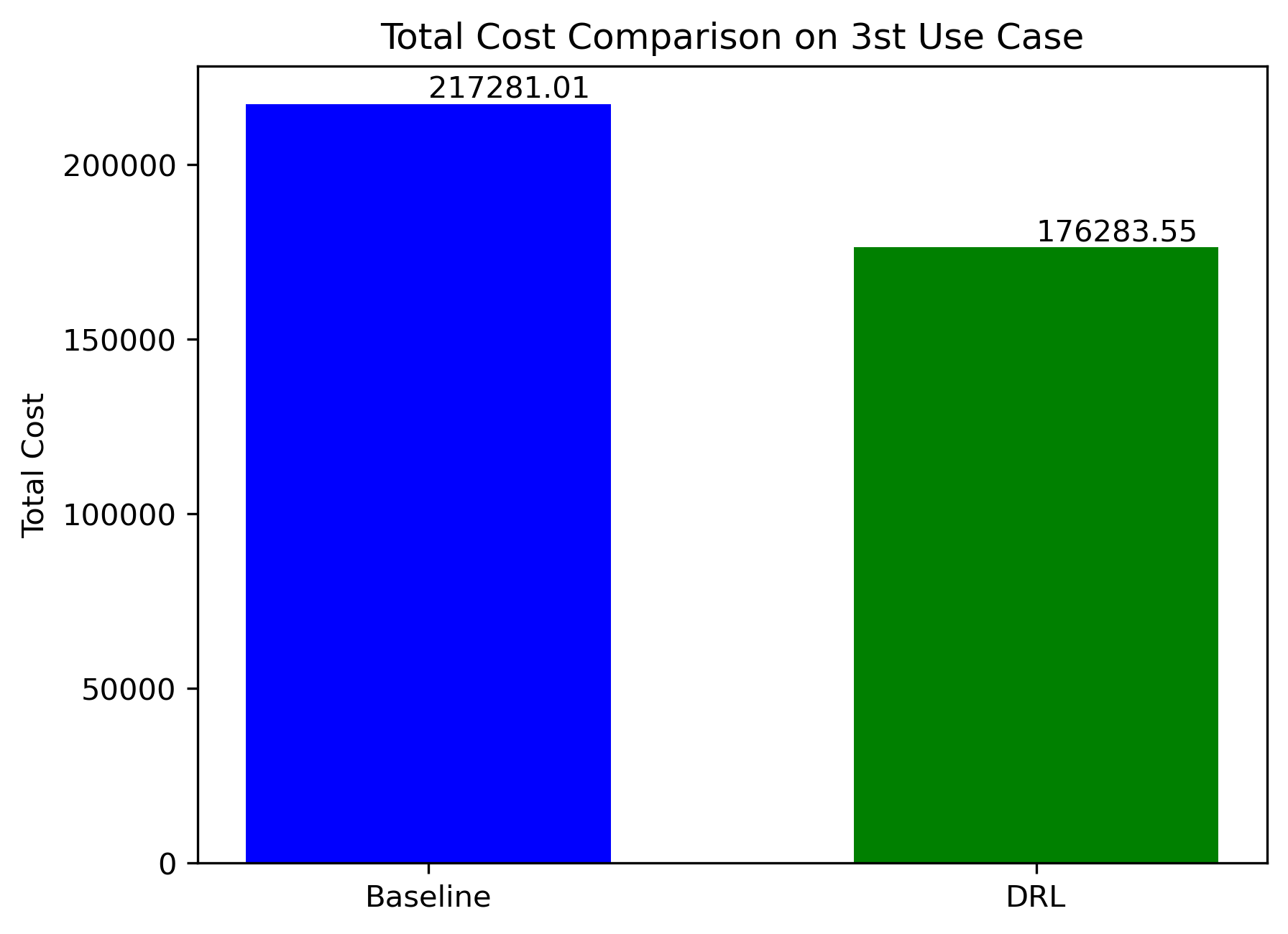}
        \caption{Third use case}
        \label{e8_res}
    \end{subfigure}
    
    \caption{Total Cost Comparisons on Different Use Cases}
    \label{fig:total_comparison}
\end{figure}

Therefore, during inference, if the number of orders in the test data differs from the graph size, the graph needs to be padded with zeros.
Then, we used the sum of LTL price of each order as the baseline to compare the performance of the DRL algorithm for all these three use cases on model 1 with epoch 100 for 1st use case and model 2 with 100 for 2nd, 3rd use cases, as shown in Figure \ref{rbcd_res}, Figure \ref{e5_res} and Figure \ref{e8_res}, where the total number of order demands for the three use cases are 171, 360, and 504, respectively.

As a result, the DRL algorithm shows a significant decrease in total transportation cost compared to baselines for the three use cases, with reductions of 23.48\%, 18.10\%, and 18.87\%, respectively. 

\begin{figure}
    \centering
    \begin{subfigure}{0.45\textwidth}
        \centering
        \includegraphics[width=\textwidth]{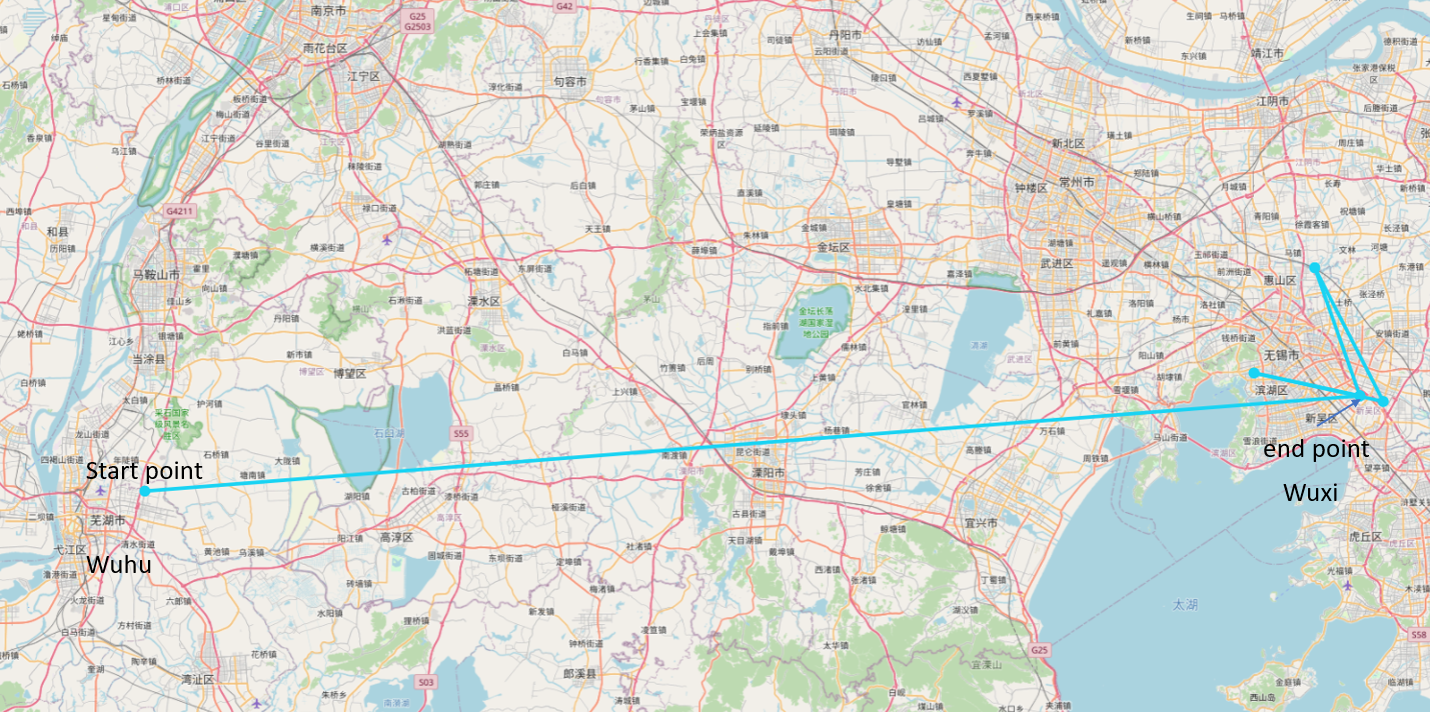}
        \caption{Visualization of the first use case}
        \label{visual_1}
    \end{subfigure}
    
    \vspace{0.2cm}
    \begin{subfigure}{0.45\textwidth}
        \centering
        \includegraphics[width=\textwidth]{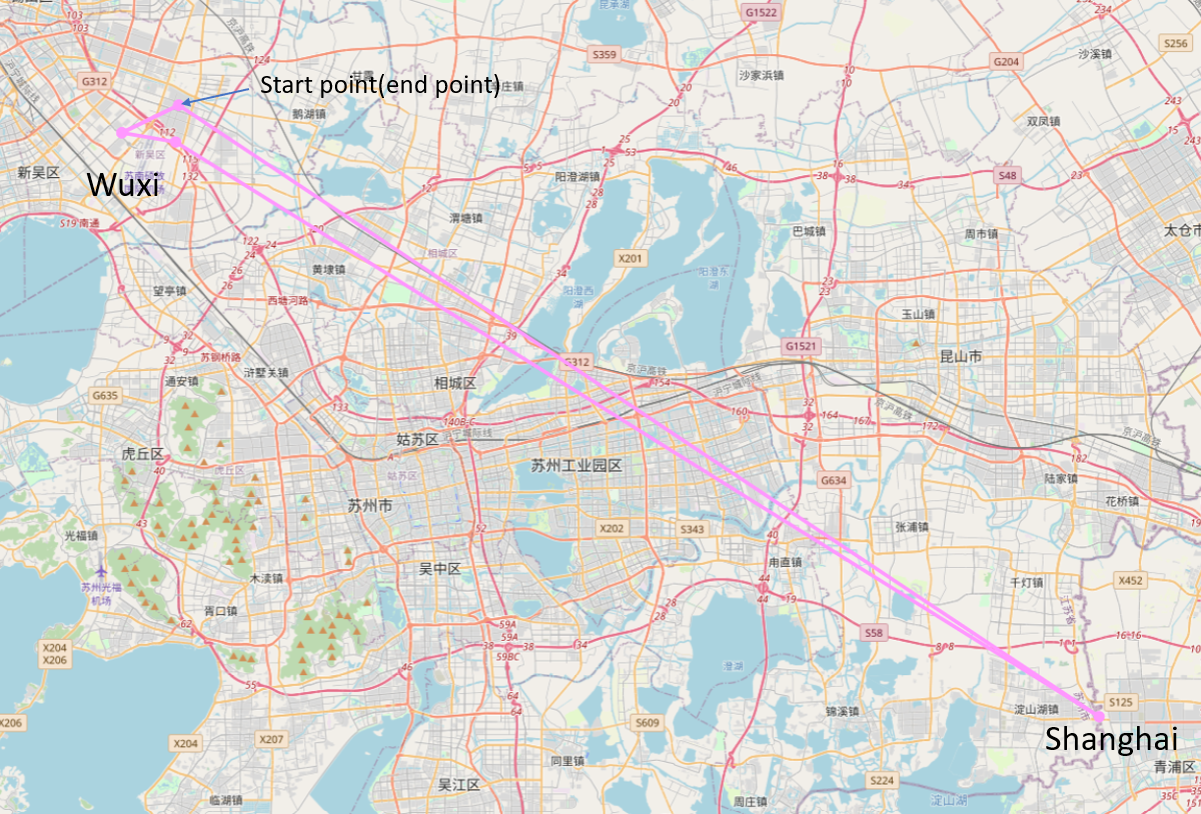}
        \caption{Visualization of the second use case}
        \label{visual_2}
    \end{subfigure}
    
     \vspace{0.2cm}
    
    \begin{subfigure}{0.45\textwidth}
        \centering
        \includegraphics[width=\textwidth]{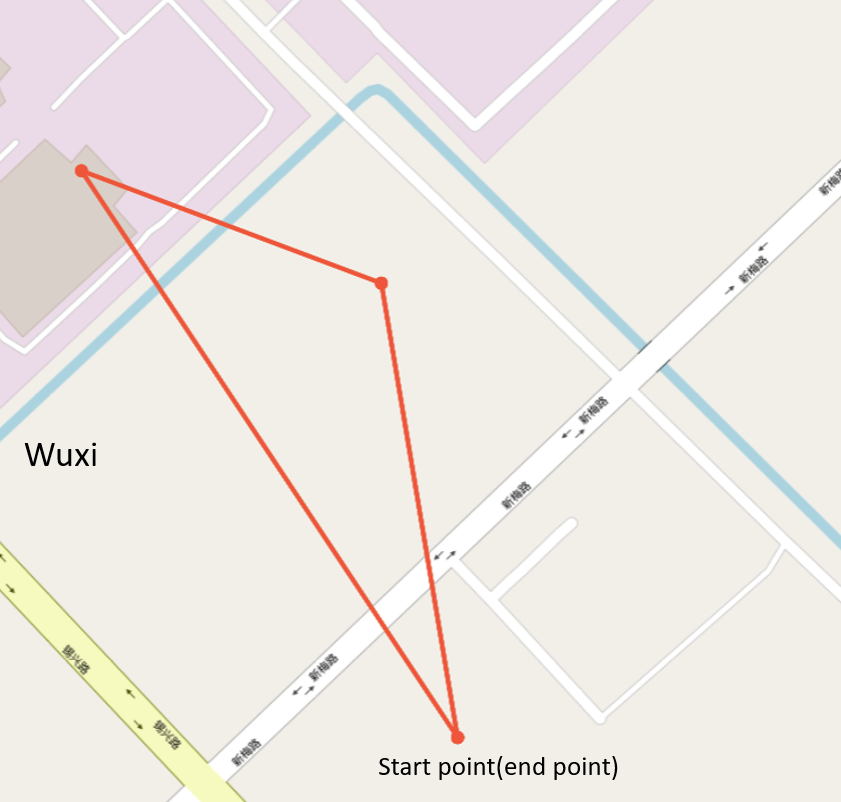}
        \caption{Visualization of the third use case}
        \label{visual_3}
    \end{subfigure}
    
    \caption{Total Cost Comparisons on Different Use Cases}
    \label{fig:total_comparison}
\end{figure}

Besides the total transportation cost savings, the detailed results consists of each newly designed EMR route, including vehicle information and the selected route for each order. Consequently, typical routes on a map from the three use cases are visualized below. In the first case shown in Figure \ref{visual_1}, the route passes 5 nodes in one trip. In the second case shown in Figure \ref{visual_2}, the route passes 4 nodes in one trip. In the third case shown in Figure \ref{visual_3}, the route passes 4 nodes in one trip. From these routes, it can be seen that the DRL algorithm is effective to generate new routes by connecting nodes together to reduce costs.

\section{Discussions}
\label{sec:prep}

This paper illustrates that the hybrid HCVRP can be solved using deep reinforcement learning by applying the reinforcement learning-based environment and algorithm, as well as analyzing the training, validation and testing results. As a result,  the inference speed of the  model is much faster than  classical optimization algorithm such as MIP and heuristic optimization. Moreover, the training process requires generating a large number of training samples with similar characteristics to the test and validation data. The results have shown that the total cost of the optimized EMR routes using reinforcement learning agents are lower than baseline results. 

In addition, since research on using reinforcement learning algorithms to handle VRP and its variants is still relatively rare compared to classic heuristic optimization algorithms, several research directions are worth exploring with reinforcement learning. For example, how to enhance the generalization capability of the agent, how to add time constraints, split frequency constraint and how to automate the number of different vehicle types selections, and how to add different capacity constraints by designing reward functions or mask functions.

Moreover, strategies about how to optimize the policy network algorithms performances (e.g., PPO) and the neural network architecture (e.g. Transformer) for more combinatorial optimization problem besides VRP are other future applied research direction in the real world.

\section{Conclusion}
\label{sec:prep}
In the field of transportation research, the Vehicle Routing Problem (VRP) is a perennially challenging issue. Experts and scholars in both industry and academia in the field of management science are constantly exploring optimization models and algorithms to effectively address routing problems. These solutions are then applied in real industrial scenarios to ultimately achieve cost optimization. The paper has illustrated the effective applications of deep reinforcement-based planning for logistics scenarios in real life, which have outperformed the baselines. The future work could focus on how to further generalize on different variations of VRP with a unified trained model so as to reduce training efforts.

\bibliographystyle{ACM-Reference-Format}
\bibliography{sample-base}










\end{document}